%% file: eccv2020submissionCR.tex
\documentclass[runningheads]{llncs}
\usepackage{graphicx}
\usepackage{tikz}
\usepackage{comment} 
\usepackage{amsmath,amssymb} 
\usepackage{color}

\input{math_commands.tex}

\usepackage{xcolor}
\usepackage{algorithm}
\usepackage{algorithmic}
\usepackage{multirow}
\usepackage{float}
\usepackage{multicol}
\usepackage{relsize}
\usepackage{amsbsy}
\usepackage{bbm}
\usepackage{soul}
\usepackage{lipsum}
\usepackage{xspace}
\usepackage{url}
\usepackage{amsbsy}
\usepackage{amsmath}
\usepackage{bbm}

\usepackage{epsfig}
\usepackage{graphicx}
\usepackage{amsmath}
\usepackage{amssymb}
\usepackage{color}
\usepackage{xcolor}
\usepackage{caption}
\usepackage{subcaption}

\usepackage{algorithm}
\usepackage{algorithmic}
\usepackage{booktabs}
\usepackage{multirow}
\usepackage{float}
\usepackage{multicol}
\usepackage{relsize}
\usepackage[]{moresize}

\newcommand{\ours}{$\text{ACL}$\xspace}
\newcommand{\fullname}{ adversarial continual learning\xspace}
\newcommand{\Fullname}{Adversarial Continual Learning\xspace}
\newcommand{\oursbold}{\textbf{$\text{ACL}$}\xspace}
\newcommand{\mnist}{$\text{5-Split MNIST}$\xspace}
\newcommand{\pmnist}{$\text{Permuted MNIST}$\xspace}
\newcommand{\cifar}{20-Split CIFAR100\xspace}
\newcommand{\mini}{$\text{20-Split miniImageNet}$\xspace}
\newcommand{\multi}{$\text{5-Datasets}$\xspace}
\newcommand{\acc}{$\text{ACC=}$\xspace}
\newcommand{\mb}{$\text{MB}$\xspace}

\newcommand{\fig}[1]{Fig.~\ref{#1}}

\begin{document}
\pagestyle{headings}
\mainmatter
\def\ECCVSubNumber{1355}  

\title{Adversarial Continual Learning} 

\titlerunning{Adversarial Continual Learning}
\author{Sayna Ebrahimi\inst{1,2}\and
Franziska Meier\inst{1}\and
Roberto Calandra\inst{1}\and\\
	Trevor Darrell\inst{2}\and
	Marcus Rohrbach\inst{1}}
\institute{\textsuperscript{1}Facebook AI Research, USA \ \ \ 	\textsuperscript{2}UC Berkeley EECS, Berkeley, CA, USA \\
\email{\{sayna,trevor\}@eecs.berkeley.edu}, 
\email{\{fmeier,rcalandra,mrf\}@fb.com}}
\authorrunning{S. Ebrahimi et al.}
\maketitle

\begin{abstract}
	Continual learning aims to learn new tasks without forgetting previously learned ones. We hypothesize that representations learned to solve each task in a sequence have a shared structure while containing some task-specific properties. We show that shared features are significantly less prone to forgetting and propose a novel hybrid continual learning framework that learns a disjoint representation for task-invariant and task-specific features required to solve a sequence of tasks. Our model combines architecture growth to prevent forgetting of task-specific skills and an experience replay approach to preserve shared skills. We demonstrate our hybrid approach is effective in avoiding forgetting and show it is superior to both architecture-based and memory-based approaches on class incrementally learning of a single dataset as well as a sequence of multiple datasets in image classification. 
	Our code is available at \url{https://github.com/facebookresearch/Adversarial-Continual-Learning}.
\end{abstract}

\section{Introduction}
Humans can learn novel tasks by augmenting core capabilities with new skills learned based on information for a specific novel task. We conjecture that they can leverage a lifetime of previous task experiences in the form of fundamental skills that are robust to different task contexts. When a new task is encountered, these generic strategies form a base set of skills upon which task-specific learning can occur. We would like artificial learning agents to have the ability to solve many tasks sequentially under different conditions by developing task-specific and task-invariant skills that enable them to quickly adapt while avoiding \textit{catastrophic forgetting} \cite{catforgetting89} using their memory. 

One line of continual learning approaches learns a single representation with a fixed capacity in which they detect important weight parameters for each task and minimize their further alteration in favor of learning new tasks. In contrast, structure-based approaches increase the capacity of the network to accommodate new tasks. However, these approaches do not scale well to a large number of tasks if they require a large amount of memory for each task. Another stream of approaches in continual learning relies on explicit or implicit experience replay by storing raw samples or training generative models, respectively. 
\begin{figure}
	\centering
	\includegraphics[width=0.7\linewidth]{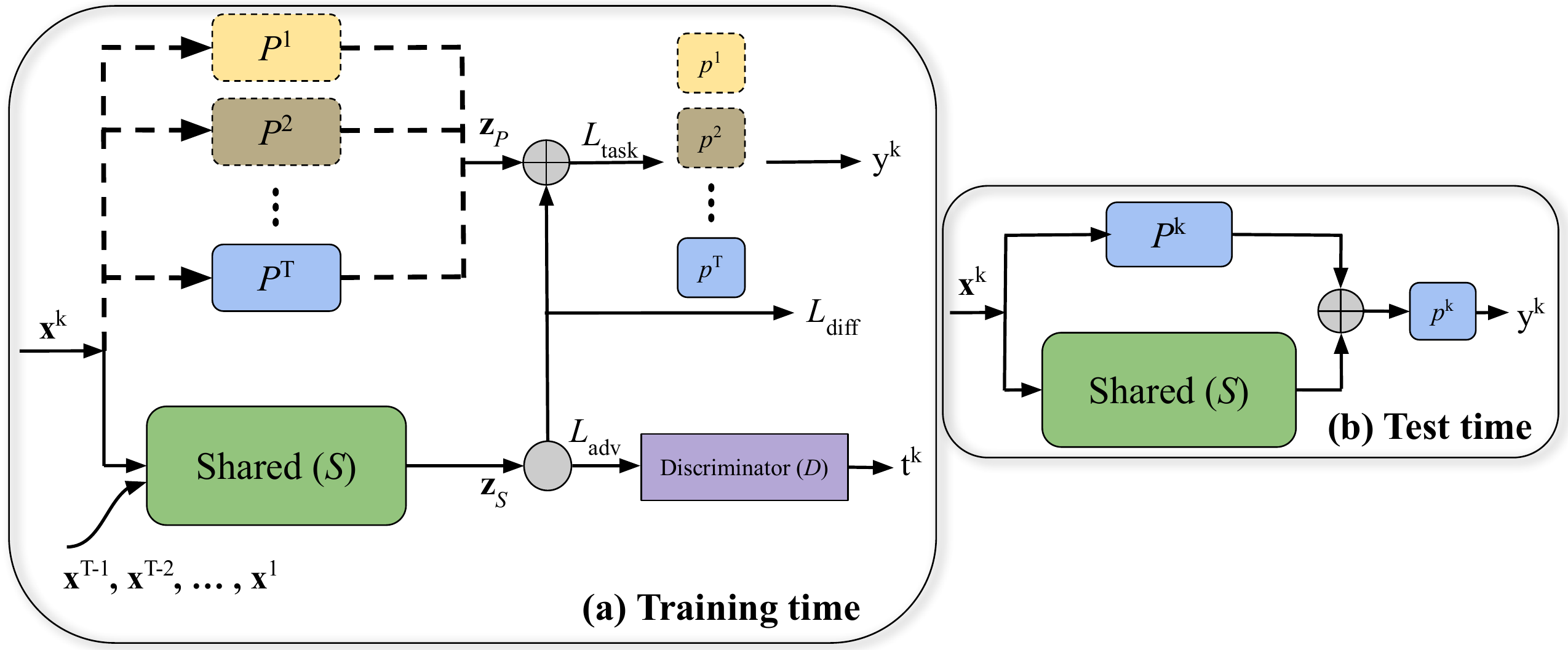}
	\caption{\footnotesize Factorizing task-specific and task-invariant features in our method (\ours) while learning $T$ sequential tasks at a time. \textit{Left:} Shows \ours at training time where the Shared module is adversarially trained with the discriminator to generate \textit{task-invariant} features ($\mathbf{z}_S$) while the discriminator attempts to predict task labels. Architecture growth occurs at the arrival of the $k^{th}$ task by adding a \textit{task-specific} modules denoted as $P^k$ and $p^k$, optimized to generate orthogonal representation $\mathbf{z}_P$ to $\mathbf{z}_S$. To prevent forgetting, 1) Private modules are stored for each task and 2) A shared module which is less prone to forgetting, yet is also retrained with experience reply with a limited number of exemplars \textit{Right:} At test time, the discriminator is removed and \ours uses the $P^k$ module for the specific task it is evaluated on.}
	\label{fig:teaser}
\end{figure}
In this paper, we propose a novel\fullname (\ours) method in which a disjoint latent space representation composed of \textit{task-specific} or \textit{private} latent space is learned for each task and a \textit{task-invariant} or \textit{shared} feature space is learned for all tasks to enhance better knowledge transfer as well as better recall of the previous tasks. The intuition behind our method is that tasks in a sequence share a part of the feature representation but also have a part of the feature representation which is task-specific. The shared features are notably less prone to forgetting and the tasks-specific features are important to retain to avoid forgetting the corresponding task. Therefore, factorizing these features separates the part of the representation that forgets from that which does not forget.
To disentangle the features associated with each task, we propose a novel adversarial learning approach to enforce the shared features to be task-invariant and employ orthogonality constraints \cite{factorized} to enforce the shared features to not appear in the task-specific space.

Once factorization is complete, minimizing forgetting in each space can be handled differently. In the task-specific latent space, due to the importance of these features in recalling the task, we freeze the private module and add a new one upon finishing learning a task. The shared module, however, is significantly less susceptible to forgetting and we only use the replay buffer mechanism in this space to the extend that factorization is not perfect, i.e., when tasks have little overlap or have high domain shift in between, using a tiny memory containing samples stored from prior tasks will help with better factorization and hence higher performance. We empirically found that unlike other memory-based methods in which performance increases by increasing the samples from prior tasks, our model requires a very tiny memory budget beyond which its performance remains constant. This alleviates the need to use old data, as in some applications it might not be possible to store a large amount of data if any at all. Instead, our approach leaves room for further use of memory, if available and need be, for architecture growth. Our approach is simple yet surprisingly powerful in not forgetting and achieves state-of-the-art results on visual continual learning benchmarks such as MNIST, CIFAR100, Permuted MNIST, miniImageNet, and a sequence of $5$ tasks. 

\section{Related Work}\label{sec:litrev}
\subsection{Continual learning}
The existing continual learning approaches can be broadly divided into three categories: memory-based, structure-based, and regularization-based methods.

\textbf{Memory-based methods:} 
Methods in this category mitigate forgetting by relying on storing previous experience explicitly or implicitly wherein the former raw samples \cite{agem,gem,rehearsal,icarl,mer} are saved into the memory for \textit{rehearsal} whereas in the latter a generative model such as a GAN \cite{genreplay} or an autoencoder \cite{fearnet} synthesizes them to perform  \textit{pseudo-rehearsal}. These methods allow for simultaneous multi-task learning on i.i.d. data which can significantly reduce forgetting.  A recent study on tiny episodic memories in CL \cite{tinymem} compared methods such as GEM~\cite{gem}, A-GEM~\cite{agem}, MER~\cite{mer}, and ER-RES \cite{tinymem}. Similar to \cite{mer}, for ER-RES they used reservoir sampling using a single pass through the data. Reservoir sampling \cite{reservoir} is a better sampling strategy for long input data compared to random selection.  In this work, we explicitly store raw samples into a very tiny memory used for replay buffer and we differ from prior work by how these stored examples are used by specific parts of our model (discriminator and shared module) to prevent forgetting in the features found to be shared across tasks. 

\textbf{Structure-based methods:} These methods exploit modularity and attempt to localize inference to a subset of the network such as columns \cite{pnn}, neurons \cite{pathnet,den}, a mask over parameters \cite{packnet,hat}. The performance on previous tasks is preserved by storing the learned module while accommodating new tasks by augmenting the network with new modules. For instance, Progressive Neural Nets (PNNs) \cite{pnn} statically grow the architecture while retaining lateral connections to previously frozen modules resulting in guaranteed zero forgetting at the price of quadratic scale in the number of parameters. \cite{den} proposed dynamically expandable networks (DEN) in which, network capacity grows according to tasks \textit{relatedness} by splitting/duplicating the most important neurons while time-stamping them so that they remain accessible and re-trainable at all time. This strategy despite introducing computational cost is inevitable in continual learning scenarios where a large number of tasks are to be learned and a fixed capacity cannot be assumed. 

\textbf{Regularization methods:} In these methods \cite{ewc,SI,mas,vcl,ucb}, the learning capacity is assumed fixed and continual learning is performed such that the change in parameters is controlled and reduced or prevented if it causes performance downgrade on prior tasks. Therefore, for parameter selection, there has to be defined a \textit{weight importance} measurement concept to prioritize parameter usage.  For instance, inspired by Bayesian learning, in elastic weight consolidation (EWC) method \cite{ewc} important parameters are those to have the highest in terms of the Fisher information matrix. HAT \cite{hat} learns an attention mask over \textit{important } parameters. Authors in \cite{ucb} used per-weight uncertainty defined in Bayesian neural networks to control the change in parameters. Despite the success gained by these methods in maximizing the usage of a fixed capacity, they are often limited by the number of tasks.  

\subsection{Adversarial learning} 
Adversarial learning has been used for different problems such as generative models~\cite{gan}, object composition~\cite{compositional}, representation learning~\cite{makhzani2015adversarial}, domain adaptation~\cite{adda}, active learning~\cite{vaal}, etc. The use of an adversarial network enables the model to train in a fully-differentiable manner by adjusting to solve the \textit{minimax} optimization problem~\cite{gan}. Adversarial learning of the latent space has been extensively researched in domain adaptation~\cite{cycada}, active learning~\cite{vaal}, and representation learning \cite{factorvae,makhzani2015adversarial}. While previous literature is concerned with the case of modeling single or multiple tasks at once, here we extend this literature by considering the case of continuous learning where multiple tasks need to be learned in a sequential manner.

\subsection{Latent Space Factorization}
In the machine learning literature, \textit{multi-view} learning, aims at constructing and/or using different views or modalities for better learning performances~\cite{blum1998combining,xu2013survey}. The approaches to tackle multi-view learning aim at either maximizing the mutual agreement on distinct views of the data or focus on obtaining a latent subspace shared by multiple views by assuming that the input views are generated from this latent subspace using Canonical correlation analysis and clustering \cite{chaudhuri2009multi}, Gaussian processes \cite{shon2006learning}, etc. Therefore, the concept of factorizing the latent space into \textit{shared} and \textit{private} parts has been extensively explored for different data modalities. Inspired by the practicality of factorized representation in handling different modalities, here we factorize the latent space learned for different tasks using adversarial learning and orthogonality constraints~\cite{factorized}.

\section{Adversarial Continual learning (ACL)}
We consider the problem of learning a sequence of $T$ data distributions denoted as $\mathcal{D}^{tr} = \{\mathcal{D}^{tr}_1, \cdots, \mathcal{D}^{tr}_T \}$, where $\mathcal{D}^{tr}_k = \{ (\mathbf{X}_i^k, \mathbf{Y}_i^k, \mathbf{T}_i^k)_{i=1}^{n_k}\}$ is the data distribution for task $k$ with $n$ sample tuples of input ($\mathbf{X}^k \in \mathcal{X}$), output label ($\mathbf{Y}^k \in \mathcal{Y}$), and task label ($\mathbf{T}^k \in \mathcal{T}$). The goal is to sequentially learn the model $f_\theta: \mathcal{X} \rightarrow \mathcal{Y}$ for each task that can map each task input to its target output while maintaining its performance on all prior tasks. We aim to achieve this by learning a disjoint latent space representation composed of a \textit{task-specific} latent space for each task and a \textit{task-invariant} feature space for all tasks to enhance better knowledge transfer as well as better catastrophic forgetting avoidance of prior knowledge. We mitigate catastrophic forgetting in each space differently. For the \textit{task-invariant} feature space, we assume a limited memory budget of $\mathcal{M}^k$ which stores $m$ samples $x_{i=1 \cdots m} \sim \mathcal{D}^{tr}_{j=1\cdots k-1}$ from every single task prior to $k$.

We begin by learning $f_\theta^k$ as a mapping from $\mathbf{X}^k$ to $\mathbf{Y}^k$. For $C$-way classification task with a cross-entropy loss, this corresponds to 

\begin{equation}\label{eq:task}
\mathcal{L}_{\text{task}}(f_\theta^{k}, \mathbf{X}^k, \mathbf{Y}^k, \mathcal{M}^k) = -\E_{(x^k, y^k)\sim (\mathbf{X}^k,\mathbf{Y}^k) \cup \mathcal{M}^k} \sum_{c=1}^{C} \mathbbm{1}_{[c=y^k]}  \log(\sigma ( f_\theta^k(x^k)))
\end{equation} 

where $\sigma$ is the softmax function and the subscript $i=\{1,\cdots,n_t\}$ is dropped for simplicity. 
In the process of learning a sequence of tasks, an ideal $f^k$ is a model that maps the inputs to two independent latent spaces where one contains the shared features among all tasks and the other remains private to each task.
In particular, we would like to disentangle the latent space into the information shared across all tasks ($\mathbf{z}_S$) and the independent
or private information of each task ($\mathbf{z}_P$) which are as distinct as possible while their concatenation followed by a task-specific head outputs the desired targets. 

To this end, we introduce a mapping called Shared ($S_{\theta_S}: \mathcal{X} \rightarrow \mathbf{z}_S)$ and train it to generate features that fool an adversarial discriminator $D$. Conversely, the adversarial discriminator ($D_{\theta_D}: \mathbf{z}_S \rightarrow \mathcal{T}$) attempts to classify the generated features by their task labels ($\mathbf{T}^{k \in \{0, \cdots, T\}}$) . This is achieved when the discriminator is trained to maximize the probability of assigning the correct task label to generated features 
while simultaneously $S$ is trained to confuse the discriminator by minimizing $\log (D( S(x^k)))$. This corresponds to the following $T$-way classification cross-entropy adversarial loss for this minimax game
\begin{equation}\label{eq:adv}
\mathcal{L}_{\text{adv}}(D, S, \mathbf{X}^k, \mathbf{T}^k, \mathcal{M}^k) = \min_S  \max_D \sum_{k=0}^{T} \mathbbm{1}_{[k=t^k]}\log \left( D \left( S \left( x^k \right) \right) \right) \,.
\end{equation}
Note that the extra label zero is associated with the `fake' task label paired with randomly generated noise features of $\mathbf{z}'_{S} \sim \mathcal{N}(\mathbf{\mu},\mathbf{\sum})$. 
In particular, we use adversarial learning in a different regime that appears in most works related to generative adversarial networks \cite{gan} such that the generative modeling of input data distributions is not utilized here because the ultimate task is to learn a discriminative representation. 

To facilitate training $S$, we use the Gradient Reversal layer \cite{DA} that optimizes the mapping to maximize the discriminator loss directly ($\mathcal{L}_{\text{task}_S} = - \mathcal{L}_{\text{D}}$). In fact, it acts as an identity function during forward propagation but negates its inputs and reverses the gradients during back propagation. 
The training for $S$ and $D$ is complete when $S$ is able to generate features that $D$ can no longer predict the correct task label for leading $\mathbf{z}_S$ to become as task-invariant as possible. The private module ($P_{\theta_P}:\mathcal{X} \rightarrow \mathbf{z}_P$), however, attempts to accommodate the task-invariant features by learning merely the features that are specific to the task in hand and do not exist in $\mathbf{z}_S$. 
We further factorize $\mathbf{z}_S$ and $\mathbf{z}_P$ by using orthogonality constraints introduced in \cite{factorized}, also known as  ``difference'' loss in the domain adaptation literature \cite{DSN}, to prevent the shared features between all tasks from appearing in the private encoded features. This corresponds to
\begin{equation}\label{eq:diff}
\mathcal{L}_{\text{diff}}(S, P, \mathbf{X}^k, \mathcal{M}^k) = \sum_{k=1}^{T} ||(S(x^k))^{\text{T}} P^k(x^k) ||^2_F,
\end{equation}
where $||\cdot||_F$ is the Frobenius norm and it is summed over the encoded features of all $P$ modules encoding samples for the current tasks and the memory.

Final output predictions for each task are then predicted using a task-specific multi-layer perceptron head which takes $\mathbf{z}_P$ concatenated with $\mathbf{z}_S$ ($\mathbf{z}_P \oplus \mathbf{z}_S$) as an input.
Taken together, these loss form the complete objective for \ours as
\begin{equation}\label{eq:aclloss}
\mathcal{L}_{\text{\ours}} = \lambda_1 \mathcal{L}_{\text{adv}} + \lambda_2 \mathcal{L}_{\text{task}} + \lambda_3 \mathcal{L}_{\text{diff}},
\end{equation}
where $\lambda_1$, $\lambda_2$, and $\lambda_3$ are regularizers to control the effect of each component. The full algorithm for \ours is given in Alg.~\ref{alg:acl}.

\subsection{Avoiding forgetting in \ours}
Catastrophic forgetting occurs when a representation learned through a sequence of tasks changes in favor of learning the current task resulting in performance downgrade on previous tasks. The main insight to our approach is decoupling the conventional \textit{single} representation learned for a sequence of tasks into two parts: a part that \textit{must not change} because it contains task-specific features without which complete performance retrieval is not possible, and a part that is \textit{less prone to change} as it contains the core structure of all tasks.

To \textit{fully prevent} catastrophic forgetting in the first part (private features), we use \textit{compact} modules that can be stored into memory. If factorization is successfully performed, the second part remains highly immune to forgetting. However, we empirically found that when disentanglement cannot be fully accomplished either because of the little overlap or large domain shift between the tasks, using a tiny replay buffer containing few samples for old data can be beneficial to retain high ACC values as well as mitigating forgetting. 

\begin{algorithm}[t]
	\caption{\Fullname (\ours))}
	\label{alg:acl}
	\scriptsize
	\begin{multicols}{2}
		\begin{algorithmic}[1]
			\FUNCTION {TRAIN($\theta_P,\theta_S,\theta_D, \mathcal{D}^{tr},\mathcal{D}^{ts}, m$)}
			\STATE Hyper-parameters: $\lambda_1,\lambda_2,\lambda_3,\alpha_S,\alpha_P,\alpha_D$
			\STATE $R \gets 0 \in \mathbb{R}^{T \times T}$
			\STATE $\mathcal{M} \gets \{\}$
			\STATE $f_\theta^{k} = f(\theta_S\oplus\theta_P)$
			\FOR {$k = 1$ \text{to T}}
			\FOR {$e = 1$ \text{to epochs}}
			\STATE Compute $\mathcal{L}_{\text{adv}}$ for $S$ using $(x,t)\in\mathcal{D}^{tr}_k \cup \mathcal{M}$  %
			\STATE Compute $\mathcal{L}_{\text{task}}$ using $(x,y) \in \mathcal{D}^{tr}_k \cup \mathcal{M}$  %
			\STATE Compute $\mathcal{L}_{\text{diff}}$ using $P^k$, $S$, and $x\in\mathcal{D}^{tr}_k$  %
			\STATE $\mathcal{L}_{\text{\ours}} = \lambda_1 \mathcal{L}_{\text{adv}} + \lambda_2 \mathcal{L}_{\text{task}} + \lambda_3 \mathcal{L}_{\text{diff}}$
			\STATE $\theta'_{S} \gets \theta_{S} - \alpha_{S} \nabla \mathcal{L}_{\text{\ours}} $      
			\STATE $\theta'_{P_t} \gets \theta_{P^k} - \alpha_{P^k} \nabla \mathcal{L}_{\text{\ours}} $  
			\STATE Compute $\mathcal{L}_{\text{adv}}$ for $D$ using $(S(x),t)$ and $(\mathbf{z}'\sim \mathcal{N}(\mathbf{\mu},\mathbf{\sum}), t=0)$  
			\STATE $\theta'_\mathrm{D} \gets \theta_{D} - \alpha_{D} \nabla \mathcal{L}_{\text{adv}}$           
			\ENDFOR
			\STATE $\mathcal{M} \gets \text{UPDATEMEMORY}(\mathcal{D}^{tr}_k,\mathcal{M}, C, m)$ 
			\STATE Store $\theta_{P^k}$   
			\STATE $f_\theta^{k} \gets f(\theta'_S\oplus\theta'_P)$
			\STATE $R_{k, \{1\cdots k\}} \gets \text{EVAL}~(f_\theta^{k},\mathcal{D}^{ts}_{\{1\cdots k\}})$ 
			\ENDFOR
			\ENDFUNCTION {}  
			
		\end{algorithmic}
		\columnbreak
		\begin{algorithmic}
			\FUNCTION {UPDATEMEMORY($\mathcal{D}^{tr}_k, \mathcal{M} , C, m$)}
			\STATE $s \gets \frac{m}{C}$ \hfill $\rhd~ s:=\#\text{~of samples per class}$
			\FOR {$c = 1$ \text{to $C$}}
			\FOR {$i = 1$ \text{to $n$}}
			\STATE $(x_i^k,y_i^k,t_i^k) \sim \mathcal{D}^{tr}_k$
			\STATE $\mathcal{M} \gets \mathcal{M} \cup (x^k,y^k,t^k)$
			\ENDFOR
			\ENDFOR 
			\STATE {\textbf{return} $\mathcal{M}$}
			\ENDFUNCTION {} 
		\end{algorithmic}
		\begin{algorithmic}
			\FUNCTION {EVAL($f_\theta^{k}, \mathcal{D}^{ts}_{\{1\cdots k\}}$)}
			\FOR {$i = 1$ \text{to $k$}}
			\STATE $R_{k,i} = \text{Accuracy}(f_\theta^{k}(x,t),y) \text{for} (x,y,t)\in \mathcal{D}^{ts}_i$
			\ENDFOR
			\STATE {\textbf{return} $R$}
			\ENDFUNCTION {} 
		\end{algorithmic}    
	\end{multicols}
\end{algorithm}

\subsection{Evaluation metrics} 
After training for each new task, we evaluate the resulting model on all prior tasks. Similar to \cite{gem,ucb}, to measure \ours performance we use ACC as the average test classification accuracy across all tasks. To measure forgetting we report backward transfer, BWT, which indicates how much learning new tasks has influenced the performance on previous tasks. While $\mathrm{BWT}<0$ directly reports \textit{catastrophic forgetting}, $\mathrm{BWT}>0$ indicates that learning new tasks has helped with the preceding tasks. 
\begin{equation}\label{eq:bwt}%
\mathrm{BWT} = \frac{1}{T-1} \sum_{i=1}^{T-1} R_{T,i} - R_{i,i} , \quad \mathrm{ACC} = \frac{1}{T} \sum_{i=1}^T R_{T,i}
\end{equation}
where $R_{n,i}$ is the test classification accuracy on task $i$ after sequentially finishing learning the $n^\mathrm{th}$ task.
We also compare methods based on the memory used either in the network architecture growth or replay buffer. Therefore, we convert them into memory size assuming numbers are $32\text{-bit}$ floating point which is equivalent to $4\text{bytes}$. 
\section{Experiments}    
In this section, we review the benchmark datasets and baselines used in our evaluation as well as the implementation details.
\subsection{ACL on Vision Benchmarks}
\noindent \textbf{Datasets:} We evaluate our approach on the commonly used benchmark datasets for $T\text{-split}$ class-incremental learning where the entire dataset is divided into $T$ disjoint susbsets or tasks. We use common image classification datasets including \textbf{\mnist} and \textbf{\pmnist} \cite{mnist}, previously used in \cite{vcl,SI,ucb}, \textbf{\cifar} \cite{cifar} used in \cite{SI,gem,agem}, and \textbf{\mini} \cite{mini} used in \cite{tinymem,proto}. We also benchmark \ours on a sequence of \textbf{\multi} including \textbf{SVHN}, \textbf{CIFAR10}, \textbf{not-MNIST}, \textbf{Fashion-MNIST} and, \textbf{MNIST} and report average performance over multiple random task orderings.  Dataset statistics are given in Table \ref{tab:datasets} in the appendix. No data augmentation of any kind has been used in our analysis. 

\noindent \textbf{Baselines:}  From the prior work, we compare with  state-of-the-art approaches in all the three categories described in Section  \ref{sec:litrev} including Elastic Weight Consolidation (EWC) \cite{ewc}, Progressive neural networks (PNNs) \cite{pnn}, and Hard Attention Mask (HAT) \cite{hat} using implementations provided by \cite{hat} unless otherwise stated.  For memory-based methods including A-GEM, GEM, and ER-RES, for \pmnist, \cifar, and \mini, we relied on the implementation provided by \cite{tinymem}, but changed the experimental setting from single to multi-epoch and %
without using $3\text{ Tasks}$ for cross-validation %
for a more fair comparison against \ours and other baselines. %
On \pmnist results for SI \cite{SI} are reported from \cite{hat}, for VCL \cite{vcl} those are obtained using their original provided code, and for Uncertainty-based CL in Bayesian framework (UCB) \cite{ucb} are directly reported from the paper. We also perform fine-tuning, and joint training. In fine-tuning (ORD-FT), an ordinary single module network without the discriminator is continuously trained without any forgetting avoidance strategy in the form of experience replay or architecture growth. In joint training with an ordinary network (ORD-JT) and our \ours setup ($\ours$-$\text{JT}$) we learn all the tasks jointly in a multitask learning fashion using the entire dataset at once which serves as the upper bound for average accuracy on all tasks, as it does not adhere to the continual learning scenario. 

\noindent \textbf{Implementation details:} For all \ours experiments except for \pmnist and \mnist  we used a reduced AlexNet \cite{alexnet} architecture as the backbone for $S$ and $P$ modules for a fair comparison with the majority of our baselines. However, \ours can be also used with more sophisticated architectures (see our code repository for implementation of \ours with reduced $\text{ResNet18}$ backbone). However, throughout this paper, we only report our results using AlexNet. The architecture in $S$ is composed of $3$ convolutional and $4$ fully-connected (FC) layers whereas $P$ is only a convolutional neural network (CNN) with similar number of layers and half-sized kernels compared to those used in $S$. The private head modules ($p$) and the discriminator are all composed of a small $3\text{-layer}$ perceptron. Due to the differences between the structure of our setup and a regular network with a single module, we used a similar CNN structure to $S$ followed by larger hidden FC layers to match the total number of parameters throughout our experiments with our baselines for fair comparisons. For \mnist and \pmnist  where baselines use a two-layer perceptron with $256$ units in each and ReLU nonlinearity, we used a two-layer perceptron of size $784\times175$ and $175\times128$ with ReLU activation in between in the shared module and a single-layer of size $784\times128$ and ReLU for each $P$. In each head, we also used an MLP with layers of size $256$ and $28$, ReLU activations, and a $14$-unit softmax layer. In all our experiments, no pre-trained model is used. We used stochastic gradient descent in a multi-epoch setting for \ours and all the baselines.

\section{Results and Discussion}
In the first set of experiments, we measure ACC, BWT, and the memory used by \ours and compare it against state-of-the-art methods with or without memory constraints on \mini. Next, we provide more insight and discussion on \ours and its component by performing an ablation study and visualizations on this dataset. In Section \ref{sec:multi}, we evaluate \ours on a more difficult continual learning setting where we sequentially train on $5$ different datasets. Finally, in section (\ref{sec:more-res}), we demonstrate the experiments on sequentially learning single datasets such as \cifar and MNIST variants. 
\subsection{\ours Performance on \mini}
Starting with \mini, we split it in $20$ tasks with $5$ classes at a time. Table \ref{tab:mini} shows our results obtained for \ours compared to several baselines. We compare \ours with HAT as a regularization based method with no experience replay memory dependency that achieves $\acc59.45\pm0.05$ with $\text{BWT=-}0.04\pm0.03\%$. Results for the memory-based methods of ER-RES and A-GEM are re(produced) by us using the implementation provided in \cite{tinymem} by applying modifications to the network architecture to match with \ours in the backbone structure as well as the number of parameters. We only include A-GEM in Table \ref{tab:mini} which is only a faster algorithm compared to its precedent GEM with identical performance. A-GEM and ER-RES use an architecture with $25.6$M parameters $(102.6\mb)$ along with storing $13$ images of size $(84\times84\times3)$ per class  $(110.1\mb)$ resulting in total memory size of $212.7\mb$. \ours is able to outperform all baselines in $\acc\mathbf{62.07\pm0.51}$, $\text{BWT=}\mathbf{0.00\pm0.00}$, using total memory of $121.6\mb$ for architecture growth $(113.1\mb)$ and storing $1$ sample per class for replay buffer $(8.5\mb)$. In our ablation study in Section \ref{sec:ablation}, we will show our performance without using replay buffer for this dataset is $\acc57.66\pm1.44$. However, \ours is able to overcome the gap by using only one image per class ($5$ per task) to achieve $\acc\mathbf{62.07\pm0.51}$ without the need to have a large buffer for old data in learning datasets like miniImagenet with diverse sets of classes. 
\begin{table*}[t]
	\caption{\footnotesize CL results on \mini measuring ACC $(\%)$, BWT $(\%)$, and Memory (\mb). (**) denotes that methods do not adhere to the continual learning setup: \ours-JT and ORD-JT serve as the upper bound for ACC for \ours/ORD networks, respectively. ($^{*}$) denotes result is re(produced) by us using the original provided code. ($\dagger$) denotes result is obtained using the re-implementation setup by \cite{hat}. BWT of Zero indicates the method is zero-forgetting guaranteed. (b) Cumulative ablation study of \ours on miniImageNet where $P$: private modules, $S$: shared module, $D$: discriminator, $\mathcal{L}_{\text{diff}}:$ orthogonality constraint, and $\text{RB}$: replay buffer memory of one sample per class. All results are averaged over $3$ runs and standard deviation is given in parentheses}
	\setlength{\tabcolsep}{2pt}
	\begin{subtable}[t]{0.5\textwidth}
		\caption{}\label{tab:mini}
		\centering
		\ssmall
		\setlength{\tabcolsep}{1pt}
		\begin{tabular}{|l|c|c|c|c|c|}
			\hline
			Method & \textbf{ACC\%} & \textbf{BWT\%} & \multicolumn{1}{c}{\begin{tabular}[c]{@{}c@{}}Arch\\ (MB)\end{tabular}} & \multicolumn{1}{c|}{\begin{tabular}[c]{@{}c@{}}Replay\\ Buffer\\ (MB)\end{tabular}} \\
			\hline
			HAT$^{*}$\cite{hat}  &  $59.45(0.05)$  & -$0.04(0.03)$ & $123.6$  & -\\     
			\hline\hline
			PNN $\dagger$ \cite{pnn}  &   $58.96(3.50) $  &   Zero    &  $588$ & -\\        
			\hline \hline
			ER-RES$^{*}$~\cite{tinymem}  & $57.32(2.56)$  &  -$11.34(2.32)$  & $102.6$ & $110.1$ \\
			A-GEM$^{*}$ \cite{agem} & $52.43(3.10)$ & -$15.23(1.45)$ & $102.6$  & $110.1$ \\ 
			\hline \hline            
			ORD-FT &  $28.76(4.56)$  & -$64.23(3.32)$  &    $37.6$  & -\\    
			ORD-JT$^{**}$ &  $69.56(0.78)$  & -  &    $5100$ & -\\    
			\ours-JT$^{**}$  & $66.89(0.32)$ & -  & $5100$ & - \\             
			\hline\hline
			\textbf{\ours (Ours)} & $\mathbf{62.07(0.51)}$ & $\mathbf{0.00(0.00)}$  & $113.1$ & $8.5$ \\  
			\hline
		\end{tabular}
	\end{subtable} \hspace{10pt}
	\begin{subtable}[t]{0.3\textwidth}
		\caption{}
		\label{tab:ablation}
		\centering
		\ssmall
		\begin{tabular}{|c|c|c|c|c|c|c|c|}
			\hline
			$\#$ & $S$ & $P$ & $D$ & $\mathcal{L}_{\text{diff}}$ & $\text{RB}$ & \textbf{ACC\%} & \textbf{BWT\%} \\ 
			\hline
			$1$ & x & & & & & $21.19(4.43)$ & -$60.10(4.14)$\\ \hline
			$2$ &  & x & & & & $29.09(5.67) $ & Zero\\\hline
			$3$ & x & & x & & & $32.82(2.71)$ & -$28.67(3.61$\\\hline
			$4$ & x & x & & x & & $49.13(3.45)$ & -$3.99(0.42)$\\\hline
			$5$ & x & x & & & & $50.15(1.41)$ & -$14.32(2.34)$\\\hline
			$6$ & x & x & & & x & $51.19(1.98)$ & -$9.12(2.98)$\\\hline
			$7$ & x & x & & x & x & $52.07(2.49)$ & -$0.01(0.01)$ \\\hline
			$8$ & x & x & x & & & $55.72(1.42)$ & -$0.12(0.34)$ \\\hline
			$9$ & x & x & x & x & & $57.66(1.44)$ & -$3.71(1.31)$ \\\hline
			$10$ & x & x & x & & x & $60.28(0.52)$ & $ 0.00(0.00)$ \\\hline
			$11$ & x & x & x & x & x & $62.07(0.51)$ & $0.00(0.00)$ \\
			\hline
		\end{tabular}
	\end{subtable}
	\vspace{-0.28in}
\end{table*}

\begin{table}[t]
	\begin{center}
		\ssmall
		\caption {\footnotesize Comparison of the effect of the replay buffer size between \ours and other baselines including A-GEM~\cite{agem}, and ER-RES~\cite{tinymem} on \mini where unlike the baselines, \ours's performance remains unaffected by the increase in number of samples stored per class as discussed in \ref{sec:ablation}. The results from this table are used to generate \fig{fig:memory} in the appendix.}
		\label{tab:memory}
		\begin{tabular}{|ll|llll|}
			\hline
			\multicolumn{2}{|l}{\textbf{Samples per class}} & \multicolumn{1}{c}{$\mathbf{1}$} & \multicolumn{1}{c}{$\mathbf{3}$} & \multicolumn{1}{c}{$\mathbf{5}$} & \multicolumn{1}{c|}{$\mathbf{13}$} \\
			\hline\hline
			\multicolumn{2}{|l}{A-GEM\cite{agem}} & $45.14(3.42)$ & $49.12(4.69)$ & $50.24(4.56)$ & $52.43(3.10)$ \\
			\multicolumn{2}{|l}{ER-RES\cite{tinymem}} & $40.21(2.68)$ & $46.87(4.51)$ & $53.45(3.45)$ & $57.32(2.56)$ \\
			\hline
			\multicolumn{1}{|c}{\multirow{2}{*}{\oursbold \textbf{(ours)}}} & ACC & $62.07(0.51)$ & $61.80(0.50)$ & $61.69(0.61)$ & $61.33(0.40)$ \\ 
			\multicolumn{1}{|c}{} & BWT & $0.00(0.00)$ & $0.01(0.00)$ & $0.01(0.00)$ & -$0.01(0.02)$ \\
			\hline
		\end{tabular}
		\vspace{-0.25in}
	\end{center}
\end{table}

\subsection{Ablation Studies on \mini}\label{sec:ablation}
We now analyze the major building blocks of our
proposed framework including the discriminator, the shared module, the private modules, replay buffer, and the \textit{difference} loss on the miniImagenet dataset. We have performed a complete cumulative ablation study for which the results are summarized in Table~\ref{tab:ablation} and are described as follows:

\noindent \textbf{Shared and private modules:} Using only a shared module without any other \ours component (ablation $\#1$ in Table \ref{tab:ablation})  yields the lowest ACC of $21.19\pm4.43$ as well as the lowest BWT performance of -$60.10\pm4.14$ while using merely private modules (ablation $\#2$) obtains a slightly better ACC of $29.05\pm5.67$ and a zero-guaranteed forgetting by definition. However, in both scenarios the ACC achieved is too low considering the random chance being $20\%$ which is due to the small size of networks used in $S$ and $P$. 

\noindent \textbf{Discriminator and orthogonality constraint ($\mathcal{L}_{\text{diff}}$):} The role of adversarial training or presence of $D$ on top of $S$ and $P$ can be seen by comparing the ablations $\#8$ and $\#5$ where in the latter $D$, as the only disentanglement mechanism, is eliminated. We observe that ACC is improved from $50.15\pm1.41$ to $55.72\pm1.42\%$ and BWT is increased from -$14.32\pm2.34$ to -$0.12\pm0.34\%$. On the other hand, the effect of orthogonality constraint as the only factorization mechanism is shown in ablation $\#4$ where the  $\mathcal{L}_{\text{diff}}$ can not improve the ACC performance, but it increases BWT form -$14.32\pm2.34$ to -$3.99\pm0.42$. Comparing ablations $\#8$ and $\#4$ shows the importance of adversarial training in factorizing the latent spaces versus orthogonality constraint if they were to be used individually. To compare the role of adversarial and diff losses in the presence of replay buffer ($\text{RB}$), we can compare $\#7$ and $\#10$ in which the $D$ and $\mathcal{L}_{\text{diff}}$ are ablated, respectively. It appears again that $D$ improves ACC more than $\mathcal{L}_{\text{diff}}$ by reaching $\acc60.28\pm0.52$ whereas $\mathcal{L}_{\text{diff}}$ can only achieve $\acc52.07\pm2.49$. However, the effect of $D$ and $\mathcal{L}_{\text{diff}}$ on BWT is nearly the same.

\noindent\textbf{Replay buffer:} Here we explore the effect of adding the smallest possible memory replay to \ours, i. e., storing one sample per class for each task. Comparing ablation $\#9$ and the most complete version of \ours ($\#11$) shows that adding this memory improves both the ACC and BWT by $4.41\%$ and $3.71\%$, respectively.
We also evaluated \ours using more samples in the memory. Table \ref{tab:memory} shows that unlike A-GEM and ER-RES approaches in which performance increases with more episodic memory, in \ours, ACC remains nearly similar to its highest performance. Being insensitive to the amount of old data is a remarkable feature of \ours, not because of the small memory it consumes, but mainly due to the fact that access to the old data might be prohibited or very limited in some real-world applications. Therefore, for a fixed allowed memory size, a method that can effectively use it for architecture growth can be considered as more practical for such applications. 

\subsection{Visualizing the effect of adversarial learning in \oursbold}\label{sec:tsne}
Here we illustrate the role of adversarial learning in factorizing the latent space learned for continually learning a sequence of tasks using the  $\text{t-distributed}$ Stochastic Neighbor Embedding (t-SNE) \cite{tsne} plots for the \mini experiment. The depicted results are obtained without using the orthogonality constraint, ($\mathcal{L}_{\text{diff}}$), to merely present the role of adversarial learning.

\begin{figure}[t]
	\centering
	\includegraphics[width=0.65\textwidth]{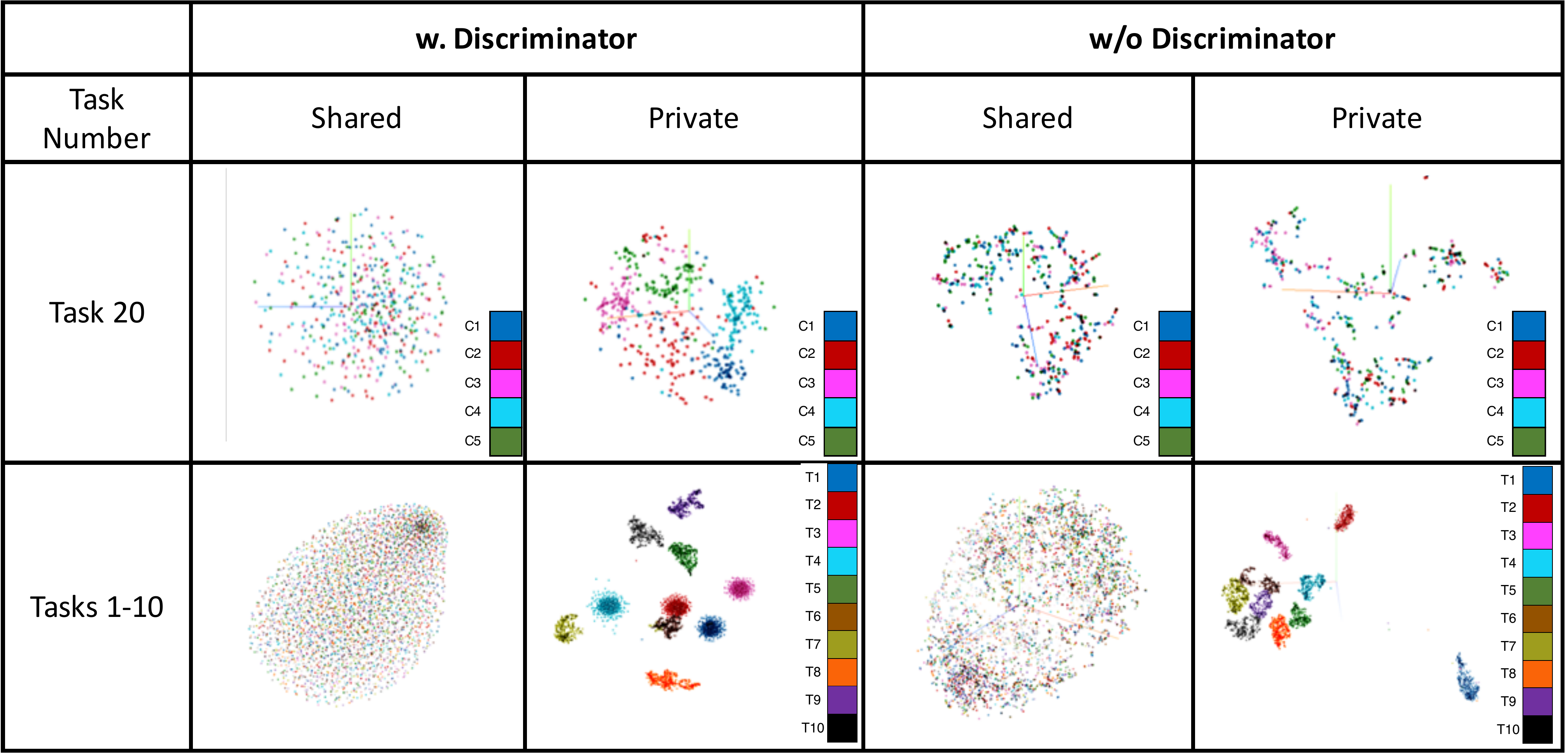}
	\caption{\footnotesize Visualizing the effect of adversarial learning on the latent space generated by shared and private modules on \mini.}   
	\label{fig:tsne}
\end{figure}
\fig{fig:tsne} visualizes the latent spaces of the shared and private modules trained with and without the discriminator. In the first row, we used the model trained on the entire sequence of \mini and evaluated it on the test-sets belonging to task $\#20$ including $100$ images for $5$ classes, a total of $500$ samples, which are color-coded with their class labels. In the second row, once again we used our final model trained on the entire sequence of \mini and tested it on the first $10$ tasks of the sequence one at a time and plotted them all in a single figure for both shared and private modules with and without the discriminator where samples are color-coded with their task labels.

We first compare the discriminator's effect on the latent space generated by the shared modules (second column). The shared modules trained with adversarial loss, consistently appear as a uniformly mixed distribution of encoded samples both for task $\#20$ and the first $10$ tasks. In contrast, in the generated features without a discriminator in the fourth column for task $\#20$, we observe a non-uniformly distributed mixture of features where small clusters can be found for some classes showing an entangled representation within each task. Similar to the pattern for task $\#20$, we observe a mixed latent space representing the shared space for the first $10$ tasks (fourth column).

The effect of the discriminator on the private modules' latent spaces is shown in the third and fifth columns where in the former, private modules trained with a discriminator, appear to be nearly successful in uncovering class labels (first row) and task labels (second row), although the final classification is yet to be happening in the private heads. As opposed to that, in the absence of the discriminator, private modules generate features as entangled as those generated by their shared module counterparts (first row). Private modules of the first $10$ tasks is not as well-clustered as the private module trained with discriminator. As we can see in the fifth column, features for task $\#6$ are spread into two parts with one close to task $\#7$ and the other next to task $\#9$.

\section{\ours Performance on a sequence of \multi}\label{sec:multi}
In this section, we present our results for continual learning of $5$ tasks using \ours in Table \ref{tab:multidatasets}. Similar to the previous experiment we look at both ACC and BWT obtained for \ours, finetuning as well as UCB as our baseline. Results for this sequence are averaged over $5$ random permutations of tasks and standard deviations are given in parenthesis. CL on a sequence of datasets has been previously performed by two regularization based approaches of UCB and HAT where UCB was shown to be superior \cite{ucb}. With this given sequence, \ours is able to outperform UCB by reaching $\acc 78.55(\pm0.29)$ and $\text{BWT=-}0.01$ using only half of the memory size and also no replay buffer. In Bayesian neural networks such as UCB, there exists double number of parameters compared to a regular model representing mean and variance of network weights.
It is very encouraging to see that \ours is not only able to continually learn on a single dataset, but also across diverse datasets.

\section{Additional Experiments}
\label{sec:more-res}
\textbf{\cifar:} In this experiment we incrementally learn $\text{CIFAR100}$ in $5$ classes at a time in $20$ tasks. As shown in Table \ref{tab:cifar}, HAT is the most competitive baseline, although it does not depend on memory and uses $27.2\mb$ to store its architecture in which it learns task-based attention maps reaching $\acc76.96\pm1.23\%$. PNN uses $74.7\mb$ to store the lateral modules to the memory and guarantees zero forgetting. %
Results for A-GEM, and ER-Reservoir are re(produced) by us using a CNN similar to our shared module architecture. We use fully connected layers with more number of neurons to compensate for the remaining number of parameters reaching $25.4\mb$ of memory. We also stored $13$ images per class ($1300$ images of size ($32\times32\times3$) in total) which requires $16.0\mb$ of memory. However, \ours achieves $\acc\mathbf{(78.08\pm1.25)}\%$ with $\text{BWT=}\mathbf{0.00\pm0.01})\%$ using only $25.1\mb$ to grow private modules with $167.2\text{K}$ parameters ($0.6\mb$) without using memory for replay buffer which is mainly due to the overuse of parameters for CIFAR100 which is considered as a relevantly `easy' dataset with all tasks (classes) sharing the same data distribution. Disentangling shared and private latent spaces, prevents \ours from using redundant parameters by only storing task-specific parameters in $P$ and $p$ modules. In fact, as opposed to other memory-based methods, instead of starting from a large network and using memory to store samples, which might not be available in practice due to confidentiality issues (\textit{e.g. } medical data), \ours uses memory to gradually add small modules to accommodate new tasks and relies on knowledge transfer through the learned shared module. The latter is what makes \ours to be  different than architecture-based methods such as PNN where the network grows by the entire \textit{column} which results in using a highly disproportionate memory to what is needed to learn a new task with. 

\begin{table*}[t]
	\caption{\footnotesize CL results on \cifar measuring ACC $(\%)$, BWT $(\%)$, and Memory (\mb). (**) denotes that methods do not adhere to the continual learning setup: \ours-JT and ORD-JT serve as the upper bound for ACC for \ours/ORD networks, respectively. ($^{*}$) denotes result is obtained by using the original provided code. ($\dagger$) denotes result is obtained using the re-implementation setup by \cite{hat}. ($^{o}$) denotes result is reported by \cite{tinymem}. BWT of Zero indicates the method is zero-forgetting guaranteed. All results are averaged over $3$ runs and standard deviation is given in parentheses.}
	
	\ssmall
	\setlength{\tabcolsep}{2pt}
	\label{tab:cifar}
	\begin{subtable}[t]{0.48\textwidth}
		\caption{\cifar }
		\centering
		\setlength{\tabcolsep}{1pt}
		\begin{tabular}{|l|c|c|c|c|}
			\hline
			Method & \textbf{ACC\%} & \textbf{BWT\%} & \multicolumn{1}{|c}{\begin{tabular}[c]{@{}c@{}}Arch\\ (MB)\end{tabular}} & \multicolumn{1}{c|}{\begin{tabular}[c]{@{}c@{}}Replay\\ Buffer\\ (MB)\end{tabular}} \\
			\hline
			HAT $^{*}$ \cite{hat}  & $76.96(1.23)$  & $0.01(0.02)$ &  $27.2$  & -\\ 
			\hline\hline
			PNN$^{\dagger}$ \cite{pnn} &  $75.25(0.04)$   &  Zero  &    $93.51$ & - \\
			\hline\hline
			A-GEM$^{o}$ \cite{agem} & $ 54.38(3.84)$    & -$21.99(4.05)$  & $25.4$  & $16$\\
			ER-RES$^{o}$~\cite{tinymem} & $66.78(0.48)$   &  -$15.01(1.11)$ &  $25.4$   & $16$ \\
			\hline \hline
			ORD-FT &  $34.71(3.36) $  & -$48.56(3.17)$    &  $27.2$ & - \\    
			ORD-JT$^{**}$ &  $78.67(0.34)$  & -  &    $764.5$  & -\\    
			\ours-JT$^{**}$ & $79.91(0.05)$ & -  &  $762.6$  & -\\             
			\hline\hline
			\textbf{\ours (Ours)} &  $\mathbf{78.08(1.25)}$ & $\mathbf{0.00(0.01)}$ & $25.1$ & - \\  
			\hline%
		\end{tabular}
	\end{subtable} \hspace{2pt}
	\ssmall	
	\begin{subtable}[t]{0.45\textwidth}
		\caption{Sequence of 5 Datasets}
		\centering
		\ssmall
		\setlength{\tabcolsep}{1pt}
		\label{tab:multidatasets}
		\begin{tabular}{|l|c|c|c|c|}
			\hline
			Method & \textbf{ACC\%} & \textbf{BWT\%} & \multicolumn{1}{|c}{\begin{tabular}[c]{@{}c@{}}Arch\\ (MB)\end{tabular}} & \multicolumn{1}{c|}{\begin{tabular}[c]{@{}c@{}}Replay\\ Buffer\\ (MB)\end{tabular}} \\
			\hline
			UCB $^{*}$ \cite{ucb}  &  $76.34(0.12)$ & $-1.34(0.04)$ & $32.8$  & - \\ 
			\hline\hline
			ORD-FT &  $27.32(2.41)$  &  -$42.12(2.57)$  & $16.5$ &  - \\    
			\hline\hline
			\textbf{\ours (Ours)} & $\mathbf{78.55(0.29)}$  &  $\mathbf{-0.01(0.15)}$ & $16.5$ & - \\  
			\hline%
		\end{tabular}
	\end{subtable}
\end{table*}

\noindent \textbf{\pmnist:} Another popular variant of the MNIST dataset in CL literature is Permuted MNIST where each task is composed of randomly permuting pixels of the entire MNIST dataset. 
To compare against values reported in prior work, we particularly report on a sequence of $T=10$ and $T=20$ tasks with ACC, BWT, and memory for \ours and baselines. To further evaluate \ours's ability in handling more tasks, we continually learned up to $40$ tasks. As shown in Table \ref{tab:pmnist} in the appendix, among the regularization-based methods, HAT achieves the highest performance of $91.6\%$ \cite{hat} using an architecture of size $1.1\mb$. Vanilla VCL improves by $7\%$ in ACC and $6.5\%$ in BWT using a K-means core-set memory size of $200$ samples per task ($
6.3\mb$) and an architecture size similar to HAT. PNN appears as a strong baseline achieving $\acc93.5\%$ with guaranteed zero forgetting. %
Finetuning (ORD-FT) and joint training (ORD-JT) results for an ordinary network, similar to EWC and HAT (a two-layer MLP with $256$ units and ReLU activations), are also reported as reference values for lowest BWT and highest achievable ACC, respectively. 
\ours achieves the highest accuracy among all baselines for both sequences of $10$ and $20$ equal to $\acc98.03\pm0.01$ and $\acc97.81\pm0.03$, and $\text{BWT=-}0.01\%$ $\text{BWT=}0\%$, respectively which shows that performance of \ours drops only by $0.2\%$ as the number of tasks doubles. \ours also remains efficient in using memory to grow the architecture compactly by adding only $55\text{K}$ parameters $(0.2\mb)$ for each task resulting in using a total of $2.4\mb$ and $5.0\mb$ when $T=10$ and $T=20$, respectively for the entire network including the shared module and the discriminator. We also observed that the performance of our model does not change as the number of tasks increases to $30$ and $40$ if each new task is accommodated with a new private module. We did not store old data and used memory only to grow the architecture by $55\text{K}$ parameters ($0.2\mb$).

\noindent \textbf{\mnist:} As the last experiment in this section, we continually learn $0\mathrm{-}9$ MNIST digits by following the conventional pattern of learning $2$ classes over $5$ sequential tasks \cite{vcl,SI,ucb}. As shown in Table \ref{tab:mnist5} in the appendix, we compare \ours with regularization-based methods with no memory dependency (EWC, HAT, UCB, Vanilla VCL) and methods relying on memory only (GEM), and VCL with K-means Core-set (VCL-C) where $40$ samples are stored per task. \ours reaches $\acc(\mathbf{99.76\pm0.03})\%$ with zero forgetting outperforming UCB with $\acc99.63\%$ which uses nearly $40\%$ more memory size. In this task, we only use architecture growth (no experience replay) where $54.3\text{K}$ private parameters are added for each task resulting in memory requirement of $1.6\mb$ to store all private modules. Our core architecture has a total number of parameters of $420.1\text{K}$. 
We also provide naive finetuning results for \ours and a regular single-module network with ($268\text{K}$) parameters ($1.1\mb$). Joint training results for the regular network (ORD-JT) is computed as $\acc99.89\pm0.01$ for \ours which requires $189.3\mb$ for the entire dataset as well as the architecture. 
\section{Conclusion}
In this work, we proposed a novel hybrid continual learning algorithm that factorizes the representation learned for a sequence of tasks into \textit{task-specific} and \textit{task-invariant} features where the former is important to be fully preserved to avoid forgetting and the latter is empirically found to be remarkably less prone to forgetting. The novelty of our work is that we use adversarial learning along with orthogonality constraints to disentangle the shared and private latent representations which results in compact private modules that can be stored into memory and hence, efficiently preventing forgetting. A tiny replay buffer, although not critical, can be also integrated into our approach if forgetting occurs in the shared module. We established a new state of the art on CL benchmark datasets.

\clearpage
\bibliographystyle{splncs04}
\bibliography{ref}
\newpage
\input{eccv2020appendixCR}

\end{document}

%% file: math_commands.tex
\usepackage{amsmath,amsfonts,bm}

\def\eqref#1{equation~\ref{#1}}

\def\1{\bm{1}}

\newcommand{\test}{\mathcal{D_{\mathrm{test}}}}

\DeclareMathAlphabet{\mathsfit}{\encodingdefault}{\sfdefault}{m}{sl}
\SetMathAlphabet{\mathsfit}{bold}{\encodingdefault}{\sfdefault}{bx}{n}

\newcommand{\E}{\mathbb{E}}



%% file: eccv2020appendixCR.tex
\onecolumn
{
	\begin{center}
		\Large 
		\textbf{Adversarial Continual Learning}
		\par
		\large
		(Supplementary Materials)\\
		\vspace{5pt}
		\small
		Sayna Ebrahimi$^{1,2}$, Franziska Meier$^{1}$, Roberto Calandra$^{1}$,\\ Trevor Darrell$^{2}$, Marcus Rohrbach$^{1}$\\
		Facebook AI Research, Menlo Park, USA\\
		UC Berkeley EECS, Berkeley, CA, USA\\
	\end{center}
} 

\par
\section{Datasets}
Table \ref{tab:datasets} shows a summary of the datasets utilized in our work. From left to right columns are given as: dataset name, total number of classes in the dataset, number of tasks, image size, number of training images per task, number of validation images per task, number of test images per task, and number of classes in each task. Statistic of \mnist and \multi experiments are given in \ref{tab:5mnist_breakdown} and \ref{tab:multi}, respectively. We did not use data augmentation for any dataset.

\begin{table*}[ht]
	\tiny
	\begin{subtable}[t]{\textwidth}
		\begin{center}
			\caption{Statistics of utilized datasets} \label{tab:datasets}
			\begin{tabular}{|l|l|l|l|l | l|l|l|} 
				\hline
				Dataset  &  $\#$Classes & $\#$Tasks & Input Size & $\#$Train/Task & $\#$Valid/Task & $\#$Test/Task &  $\#$Class/Task \\  
				\hline \hline
				\mnist \cite{mnist} & $10$ & $5$ & $1\times28\times28$  & $10$ & see Tab.  \ref{tab:5mnist_breakdown} & see Tab. \ref{tab:5mnist_breakdown} & $2$ \\ 
				\pmnist \cite{pmnist} & $10$ & $10/20/30/40$ &  $1\times28\times28$ & $51,000$ & $9,000$ & $10,000$ & $10$ \\
				\cifar \cite{cifar} & $100$ & $20$ & $3\times32\times32$ & $2,125$ & $375$ & $500$  & $5$ \\ 
				\mini \cite{imagenet}& $100$ & $20$ &$3\times84\times84$ & $2,125$ & $375$ & $500$  & $5$\\ 
				\hline\hline
				\multi& $5\times10$ & $5$ &$3\times32\times32$ & see Tab. \ref{tab:multi} & see Tab. \ref{tab:multi} & see Tab. \ref{tab:multi} & $10$  \\ 
				[0.5ex]
				\hline
			\end{tabular}
		\end{center}
	\end{subtable} \\
	\begin{subtable}[t]{\textwidth}
		\begin{center}
			\caption{Number of training, validation, and test samples per task for \mnist } \label{tab:5mnist_breakdown}
			\begin{tabular}{@{}|l|c|l|c|l|cl|c|l|c|l|@{}}
				\hline
				Task number           & \multicolumn{2}{c|}{\begin{tabular}[c]{@{}c@{}}T = 1\\  (0,1)\end{tabular}} & \multicolumn{2}{c|}{\begin{tabular}[c]{@{}c@{}}T=2\\ (2,3)\end{tabular}} & \multicolumn{2}{c|}{\begin{tabular}[c]{@{}c@{}}T=3\\ (4,5)\end{tabular}} & \multicolumn{2}{c|}{\begin{tabular}[c]{@{}c@{}}T=4\\ (6,7)\end{tabular}} & \multicolumn{2}{c|}{\begin{tabular}[c]{@{}c@{}}T=5\\ (8,9)\end{tabular}} \\ \hline \hline
				\# Training samples   & \multicolumn{2}{c|}{10766}                                                 & \multicolumn{2}{c|}{10276}                                               & \multicolumn{2}{c|}{9574}                                               & \multicolumn{2}{c|}{10356}                                               & \multicolumn{2}{c|}{10030}                                               \\
				\# Validation samples & \multicolumn{2}{c|}{1899}                                                  & \multicolumn{2}{c|}{1813}                                                & \multicolumn{2}{c|}{1689}                                               & \multicolumn{2}{c|}{1827}                                                & \multicolumn{2}{c|}{1770}                                                \\
				\# Test samples       & \multicolumn{2}{c|}{2115}                                                  & \multicolumn{2}{c|}{2042}                                                & \multicolumn{2}{c|}{1874}                                               & \multicolumn{2}{c|}{1986}                                                & \multicolumn{2}{c|}{1983}                                                \\ 
				[0.5ex]
				\hline
			\end{tabular}
		\end{center}
	\end{subtable}
	
	\begin{subtable}[t]{\textwidth}
		\begin{center}
			\caption{Statistics of utilized datasets in \multi. MNIST, notMNIST, and Fashion MNIST are padded with $0$ to become $32\times32$ and have $3$ channels. } \label{tab:multi}
			\begin{tabular}{|l|l|l|l|l|l|}
				\hline
				Dataset              & \multicolumn{1}{l|}{MNIST}   & \multicolumn{1}{l|}{notMNIST} & \multicolumn{1}{l|}{Fashion MNIST} & \multicolumn{1}{l|}{CIFAR10} & \multicolumn{1}{l|}{SVHN}    \\ \hline\hline
				\multicolumn{1}{|l|}{\# Training samples}   & \multicolumn{1}{c|}{51,000} & \multicolumn{1}{c|}{15,526}  & \multicolumn{1}{c|}{9,574}        & \multicolumn{1}{c|}{42,500} & \multicolumn{1}{c|}{62,269} \\
				\multicolumn{1}{|l|}{\# Validation samples} & \multicolumn{1}{c|}{9,000}  & \multicolumn{1}{c|}{2,739}   & \multicolumn{1}{c|}{1,689}        & \multicolumn{1}{c|}{7,500}  & \multicolumn{1}{c|}{10,988} \\
				\multicolumn{1}{|l|}{\# Test samples}       & \multicolumn{1}{c|}{10,000} & \multicolumn{1}{c|}{459}     & \multicolumn{1}{c|}{1,874}        & \multicolumn{1}{c|}{10,000} & \multicolumn{1}{c|}{26,032} \\ 
				[0.5ex]
				\hline
			\end{tabular}
		\end{center}
	\end{subtable}
\end{table*}

\section{Results on \pmnist and \mnist}
Table \ref{tab:pmnist} and \ref{tab:mnist5} show results obtained for \pmnist and \mnist experiments described in section \ref{sec:more-res}, respectively. 

\begin{table}[ht]
	\ssmall
	\begin{center}
		\caption{\footnotesize CL results on \pmnist.
			measuring ACC $(\%)$, BWT $(\%)$, and Memory (\mb). (**) denotes that methods do not adhere to the continual learning setup: \ours-JT and ORD-JT serve as the upper bound for ACC for \ours/ORD networks, respectively. 
			($^{*}$) denotes result is obtained by using the original provided code. ($\ddagger$) denotes result reported from original work. ($^{o}$) denotes the results reported by \cite{tinymem} and ($^{oo}$) denotes results are reported by \cite{hat}; T shows the number of tasks. Note the difference between BWT of Zero and $0.00$ where the former indicates the method is zero-forgetting guaranteed by definition and the latter is computed using Eq. \ref{eq:bwt}. All results are averaged over $3$ runs, the standard deviation is provided in parenthesis.}
		\label{tab:pmnist}
		\begin{tabular}{|l|c|c|c|c|}
			\hline
			Method & \textbf{ACC\%} & \textbf{BWT\%} & \multicolumn{1}{|c}{\begin{tabular}[c]{@{}c@{}}Arch\\ (MB)\end{tabular}} & \multicolumn{1}{c|}{\begin{tabular}[c]{@{}c@{}}Replay\\ Buffer\\ (MB)\end{tabular}} \\
			\hline
			EWC$^{oo}$ \cite{ewc} ($\text{T=10}$) & $88.2$ & -  & $1.1 $ &- \\ 
			HAT$\ddagger$ \cite{hat} ($\text{T=10}$) & $97.4$ & - & $2.8 $ &- \\ 
			UCB$\ddagger$ \cite{ucb} ($\text{T=10}$)& $91.44(0.04)$  & -$0.38(0.02)$ &   $2.2 $&  - \\ 
			VCL$^{*}$\cite{vcl} ($\text{T=10}$)  & $88.80(0.23)$  &  -$7.90(0.23) $ &  $1.1 $  &- \\
			VCL-C$^{*}$ \cite{vcl}($\text{T=10}$) & $95.79(0.10)$ & -$1.38(0.12)$ & $1.1 $  & $6.3$ \\ 
			\hline\hline
			PNN$^{o}$ \cite{pnn} ($\text{T=20}$) & $93.5(0.07)$   & Zero  &  N/A & - \\
			\hline \hline
			ORD-FT  ($\text{T=10}$)  &  $44.91(6.61)$ &  -$53.69(1.91)$ &  $1.1 $ &- \\    
			ORD-JT$^{**}$ ($\text{T=10}$) & $96.03(0.02)$  & -  &  $189.3 $ &- \\    
			\ours-JT$^{**}$ ($\text{T=10}$) & $98.45(0.02)$  &  - & $194.4 $&  - \\            
			\hline\hline
			\textbf{\ours(Ours)} ($\text{T=10}$) & $\mathbf{98.03(0.01)}$  & -$0.01(0.01)$  & $2.4$ & -\\      
			\textbf{\ours(Ours)} ($\text{T=20}$)& $\mathbf{97.81(0.03)}$  & $0.00(0.00)$ & $5.0$ & - \\
			\textbf{\ours(Ours)} ($\text{T=30}$)& $\mathbf{97.81(0.03)}$  & $0.00(0.00)$ & $7.2$ & - \\
			\textbf{\ours(Ours)} ($\text{T=40}$)& $\mathbf{97.80(0.02)}$  & $0.00(0.00)$ & $9.4$ & - \\
			\hline
		\end{tabular}
	\end{center}
\end{table}

\section{Effect of memory size on \ours}
\begin{figure}[ht]
	\centering
	\includegraphics[width=0.65\linewidth]{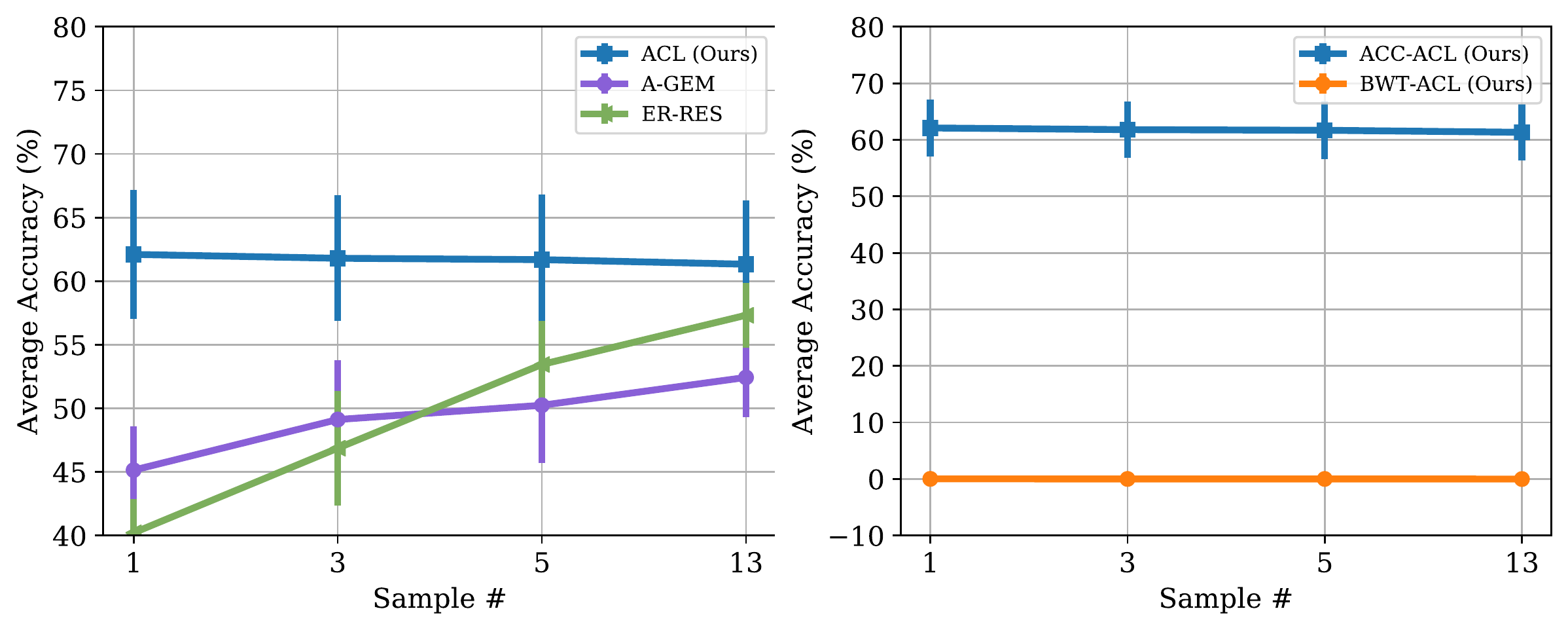}
	\caption{\footnotesize \textit{Left:} Comparing the replay buffer effect on ACC on \mini achieved by \ours against A-GEM~\cite{agem} and ER-RES~\cite{tinymem} when using $1$, $3$, $5$, and $13$ samples per classes within each task discussed in \ref{sec:ablation}. \textit{Right:} Insensitivity of ACC and BWT to replay buffer in \ours. Best viewed in color.}
	\label{fig:memory}
\end{figure}

\fig{fig:memory} shows the effect of memory size on \ours and memory-dependent baselines when $1$, $3$, $5$, and $13$ images per class are used during training where in the left it illustrates the memory effect for \ours and memory-dependent baselines. We also show how memory affects the BWT in \ours in \fig{fig:memory} (right) which follows the same pattern as we observed for ACC. Numbers used to plot this figure with their standard deviation are given in Table \ref{tab:memory}.

\begin{table}[ht]
	\begin{center}
		\caption{\small Class Incremental Learning on \mnist.
			measuring ACC $(\%)$, BWT $(\%)$, and Memory (\mb). (**) denotes that methods do not adhere to the continual learning setup: \ours-JT and ORD-JT serve as the upper bound for ACC for \ours/ORD networks, respectively. ($^{*}$) denotes result is obtained by using the original provided code. ($\ddagger$) denotes result reported from original work and ($\dagger\dagger$) denotes results are reported by \cite{ucb};
			All results are averaged over $3$ runs, the standard deviation is provided in parenthesis}
		\centering
		\ssmall
		\setlength{\tabcolsep}{1pt}
		\begin{tabular}{|l|c|c|c|c|}
			\hline
			Method & \textbf{ACC\%} & \textbf{BWT\%} & \multicolumn{1}{c}{\begin{tabular}[c]{@{}c@{}}Arch\\ (MB)\end{tabular}} & \multicolumn{1}{c|}{\begin{tabular}[c]{@{}c@{}}Replay\\ Buffer\\ (MB)\end{tabular}} \\
			\hline
			EWC $\dagger\dagger$ \cite{ewc} & $95.78(0.35)$ & -$4.20(0.21)$  &  $1.1$ & -\\ 
			HAT $\dagger\dagger$ \cite{hat}  & $99.59(0.01)$ & $0.00(0.04)$ &  $1.1$ & - \\ 
			UCB $\ddagger$ \cite{ucb} & $99.63(0.02)$  & $0.00(0.00)$ &  $2.2$ & - \\ 
			VCL $^{*}$\cite{vcl}   & $95.97(1.03)$  &  -$4.62(1.28) $ &   $1.1$ & -\\
			\hline\hline
			GEM$^{*}$ \cite{gem}   & $94.34(0.82)$  & -$2.01(0.05)$  &  $6.5$ & $0.63$ \\             
			VCL-C $^{*}$ \cite{vcl}& $93.6(0.20)$ & -$3.10(0.20)$ &  $1.7$ & $0.63$ \\ 
			\hline \hline
			ORD-FT &   $65.96(3.53)$ & -$40.15(4.27) $  & $1.1$& - \\
			ORD-JT$^{**}$ &   $99.88(0.02)$  & -   &     $189.3$ & - \\    
			\ours-JT$^{**}$ (Ours) &  $99.89(0.01)$ & -  &  $190.8$ & -  \\ 
			\hline\hline
			\textbf{\ours (Ours)} & $\mathbf{99.76(0.03)}$ & $0.01(0.01)$   & $1.6 $ & -\\  
			\hline
		\end{tabular}
		\label{tab:mnist5}
	\end{center}
\end{table}

\section{Intransigence Measure}
Table \ref{tab:intransigence} shows our results for Intransigence ($I$) measure introduced in \cite{riemannian} and computed for the $k$-th task as below:
\begin{equation}
I_k = a_k^* - a_{k,k}
\end{equation}
where $a_k^*$ is the accuracy on the $k\text{-th}$ task using a reference model trained with all the seen datasets up to task $k$ and $a_{k,k}$ denotes the accuracy on the $k\text{-th}$ task when trained up to task $k$ in an incremental manner. $I_k \in [-1, 1]$. 

Table \ref{tab:intransigence} shows $I$ evaluated at the end of tasks sequence for \ours and the strongest baselines (HAT \cite{hat} and ER-RES) in \mnist, \pmnist, \cifar, and \mini experiments. We found \ours achieves the lowest $I$ (lower the better) value compared to baselines, always negative, which means in \ours, learning tasks up to a specific task in hand increases the model's ability to learn new ones which is consistent with our idea of proposing a shared latent space for better initialization for new tasks.

\begin{table}[ht]
	\tiny
	\begin{center}
		\caption{\small Intransigence measure results obtained for \mnist, \pmnist, \cifar, and \mini experiments in HAT \cite{hat}, ER-RES, and \ours. $I_k$ denotes the measure computed at the end of the $k$-th task.}
		\label{tab:intransigence}
		\begin{tabular}{|l|c|c|c|c|}
			\hline
			& \mnist ($I_5$) & \pmnist ($I_{10}$) & \cifar ($I_{20}$) & \mini ($I_{20}$) \\ 
			\hline
			HAT \cite{hat} & -$0.0001$ & -$0.01$ & $0.1$ & $0.1$ \\
			ER-RES & $0.002$ & $0.001$ & -$0.002$ & -$0.02$ \\
			\hline \hline
			\oursbold (\textbf{Ours}) & -$\mathbf{0.5}$ & -$\mathbf{0.04}$ & -$\mathbf{0.03}$ & -$\mathbf{0.2}$ \\
			\hline
		\end{tabular}
	\end{center}
\end{table}